\documentclass[runningheads]{llncs}
\usepackage{cite}
\usepackage[T1]{fontenc}
\usepackage{graphicx}
\usepackage{algorithm}  
\usepackage{algpseudocode}  
\usepackage{amsmath}  
\usepackage{makecell}
\usepackage{multirow}
\usepackage{xcolor}
\usepackage{marvosym}
\usepackage{hyperref}
\begin{document}
\title{PromptAttack: Prompt-based Attack for Language Models via Gradient Search\thanks{This research is supported by the National Natural Science Foundation of China (No. 62106105) and the National Key R\&D Program of China (No. 2021YFB3100700).\\
\Letter\ Corresponding author.}}
\titlerunning{PromptAttack}
% % If the paper title is too long for the running head, you can set
% % an abbreviated paper title here
%\orcidID{0000-1111-2222-3333}
% \author{Yundi Shi\inst{1} \and
% Piji Li\inst{1}\textsuperscript{(\Letter)} \and
% Changchun Yin\inst{1} \and
% Zhaoyang Han\inst{1} \and
% Lu Zhou\inst{1} \and
% Zhe Liu\inst{1} }
% %
\author{Yundi Shi \and
Piji Li\textsuperscript{\Letter} \and
Changchun Yin \and
Zhaoyang Han \and
Lu Zhou \and
Zhe Liu }

\authorrunning{Shi et al.}

\institute{Nanjing University of Aeronautics and Astronautics, Nanjing, Jiangsu, China
\email{\{shiyundi,pjli,ycc0801,sunrisehan,lu.zhou,zhe.liu\}@nuaa.edu.cn} \\}

\maketitle              % typeset the header of the contribution%
\begin{abstract}
As the pre-trained language models (PLMs) continue to grow, so do the hardware and data requirements for fine-tuning PLMs. Therefore, the researchers have come up with a lighter method called \textit{Prompt Learning}. However, during the investigations, we observe that the prompt learning methods are vulnerable and can easily be attacked by some illegally constructed prompts, resulting in classification errors, and serious security problems for PLMs. Most of the current research ignores the security issue of prompt-based methods.
%, but we found that the specific template part of \textit{prompt} is vulnerable.
Therefore, in this paper, we propose a malicious prompt template construction method (\textbf{PromptAttack}) to probe the security performance of PLMs.
Several unfriendly template construction approaches are investigated to guide the model to misclassify the task. Extensive experiments on three datasets and three PLMs prove the effectiveness of our proposed approach PromptAttack. We also conduct experiments to verify that our method is applicable in few-shot scenarios.

\end{abstract}
\section{Introduction}
The emergence of pre-trained language models (PLMs) has facilitated the development of Natural Language Processing (NLP)~\cite{han2021pre}. The research approach based on PLMs is usually in a ``pre-train $\rightarrow$ fine-tune'' paradigm. In the pre-training stage, the models are trained using a large amount of data. In the fine-tuning stage, the PLMs are tuned by small datasets (compared with the data size in the pre-training stage) collected for different downstream tasks. The method achieves good results on many NLP tasks.
%and then the models can adapt various downstream tasks.

%~\cite{tam2021improving}
%,raffel2019exploring,radford2018improving,}.

However, as the size of the pre-trained models continues to increase, the hardware and data requirements for fine-tuning PLMs are also increasing~\cite{liu2021pre}. At the same time, diverse downstream tasks also lead to complex model design. The researchers propose a method for setting fill-in-the-blank templates for downstream tasks. The purpose is to make the downstream tasks use the model as consistently as possible with the tasks in the pre-training phase without fine-tuning all the parameters. It has gradually evolved into prompt learning~\cite{liu2021pre}, a lighter approach to solve NLP tasks.
%\vspace{-0.5cm}
\begin{figure}[!t]
\centering
\includegraphics[width=\textwidth]{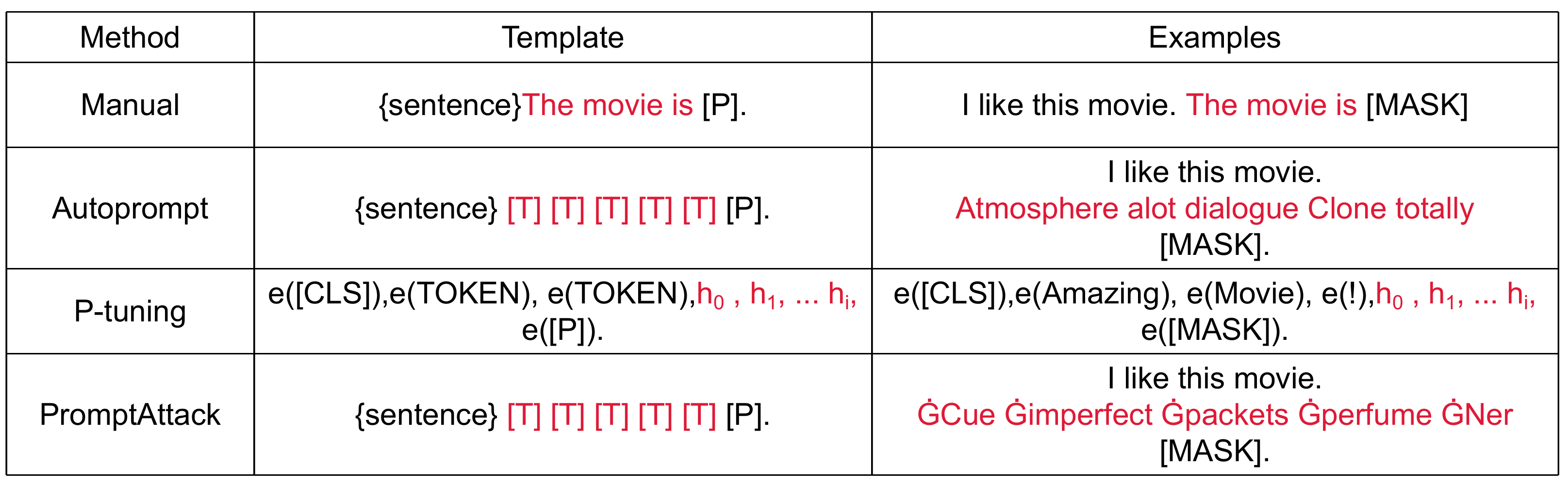}
\caption{Prompts of different approaches.} \label{fig0}
\vspace{-5mm}
\end{figure}

The prompt-based model consists of three steps: First, we need a PLM with the masked pre-training task. Second, we need to construct a template in cloze form. For example, in the sentiment classification task, for the sentence ``This is the best movie I've ever watched.'', we can create the template like ``The movie is [MASK]'', and then use PLMs to predict which emotional word (e.g.``great'', ``bad'') should be filled into the masked position. At last, the answer predicted by PLMs is converted into the real label. Words such as ``great'' and ``wonderful'' should correspond to positive, and words such as ``terrible'' and ``bad'' should correspond to negative. We call this map a verbalizer. Initially, the templates are set manually. However, if we want to design an efficient template, we require numerous experiments, much relevant knowledge, and high cost. So researchers begin to explore how to build templates automatically. In the first three rows in Fig.\ref{fig0}, we show the examples of the manual template (Manual), the automatic discrete template (Autoprompt~\cite{schick2020automatically}), and the automatic continuous template (P-tuning~\cite{liu2021gpt}).

In recent years, text analysis and understanding based on deep learning have become the core technology of various NLP applications. Despite its popularity and excellent performance, studies have shown that the models are \textbf{vulnerable to malicious attacks}. Considering its increasing application in many real-world security-sensitive tasks, the vulnerability has caused great concern and high attention to the models' security. Therefore, safety has gradually become a new research hotspot, and more and more researchers focus on attack and defense. Although the prompt has shown excellent performance in various tasks, including text understanding and text generation, its \textbf{vulnerability and security} has not been comprehensively and deeply evaluated. Deploying such models in real systems may have large security risks, and using these models without sufficient awareness of potential security risks may lead to serious consequences. Therefore, it is very necessary to conduct in-depth research on the security issues of the prompt-based model. Recently, adversarial attacks and backdoor attacks have become the primary threat in the field of AI security. We believe that the two attack methods are also applicable to the prompt-based model. Besides, we think that the attack can also target the template which is the unique part of the prompt. Many papers have verified the choice of the template has a significant impact on the model prediction results. Therefore, in theory, if the template we build is malicious, the accuracy of the model prediction can be greatly reduced.
%~\cite{logan2021cutting}

According to the above analysis, we propose an attack method  \textbf{PromptAttack} to construct malicious templates by automatically searching for discrete tokens. First, our label map is automatically selected. We use the logical classifier to find the relationship between the word and the label, and select the \textit{top-k} words with the highest correlation value as the corresponding words of the label. Second, we find tokens according to a variant of the gradient search strategy and select the \textit{top-k} tokens, which meet our requirements, as candidate words. Then, we replace the token of the existing templates on the principle of random replacement or beam search. We calculate the prediction accuracy of each template on the train dataset and select the one that minimizes the prediction accuracy as the final template. Finally, we consider the stealth of the attack, i.e. the readability of the constructed template, so we choose to utilize GPT-2 to generate the template with lower perplexity. In the last row of Fig.\ref{fig0}, we depict an example of templates created by our proposed attack method. The contributions of our paper are as follows:

(1) We consider that the prompt still has the risk of being attacked, and almost no other papers have considered this problem.

(2) Our attack method PromptAttack is to attack the prompt-specific template.

(3) We evaluate the attack method on three datasets and three pre-trained models, and the experimental results prove the effectiveness of PromptAttack.

(4) Our attack method is still effective in few-shot cases.

\section{Related work}
\subsection{Prompt Learning}In the recent two years, prompt learning has made great progress, from discrete prompts to continuous prompts, from text-only prompts to multimodal prompts, and from white-box prompts to black-box prompts. Schick et al.~\cite{schick2020exploiting} proposed the Pattern-Exploiting Training (PET) method. The template and verbalizer for this method are manually defined. Schick~\cite{schick2020automatically} also proposed the method of using the Likelihood Ratio to automatically search verbalizer. Shin et al.~\cite{shin2020autoprompt} designed a discrete prompt named Autoprompt, and in their method, both the template and the verbalizer are automatically constructed. The \textit{prefix-tuning} proposed by Li et al.~\cite{li2021prefix}, and the \textit{P-tuning} proposed by Liu et al.~\cite{liu2021gpt} are typical methods for constructing continuous prompts. \textit{CPT} by Yao et al.~\cite{yao2021cpt}, \textit{CLIP} by Radford et al.~\cite{radford2021learning}, and \textit{CoOp} by Zhou et al.~\cite{zhou2021learning} are all representative methods of multimodal prompt.

\subsection{Attack methods}
Attack methods are mainly divided into two forms: adversarial attack and backdoor attack. The adversarial attack occurs in the model testing phase. Papernot et al.~\cite{papernot2016crafting} first pointed out that attackers can generate adversarial examples by adding noise to the text, which may make the classifier misclassify. Ebrahimi et al.~\cite{ebrahimi2017hotflip} proposed an attack method \textit{HotFlip}, which can generate adversarial examples in a white-box situation. Li et al.~\cite{li2018textbugger} proposed the \textit{TEXTBUGGER} method, which can generate adversarial examples in both white-box and black-box situations.The backdoor attack occurs in the model training phase. The attacker injects poisoned samples into the train dataset, thereby embedding backdoor triggers in the trained deep learning model. Backdoor attacks were first proposed by Gu et al.~\cite{gu2019badnets} and further exploited on NLP models (Kurita et al.~\cite{kurita2020weight}). For the attack of the prompt method, there are only the \textit{AToP} and \textit{BToP} methods proposed by Xu et al.~\cite{xu2022exploring}.

\section{Methodology}
\subsection{Overview} 
\begin{figure}[!t]
\centering
\includegraphics[width=\textwidth]{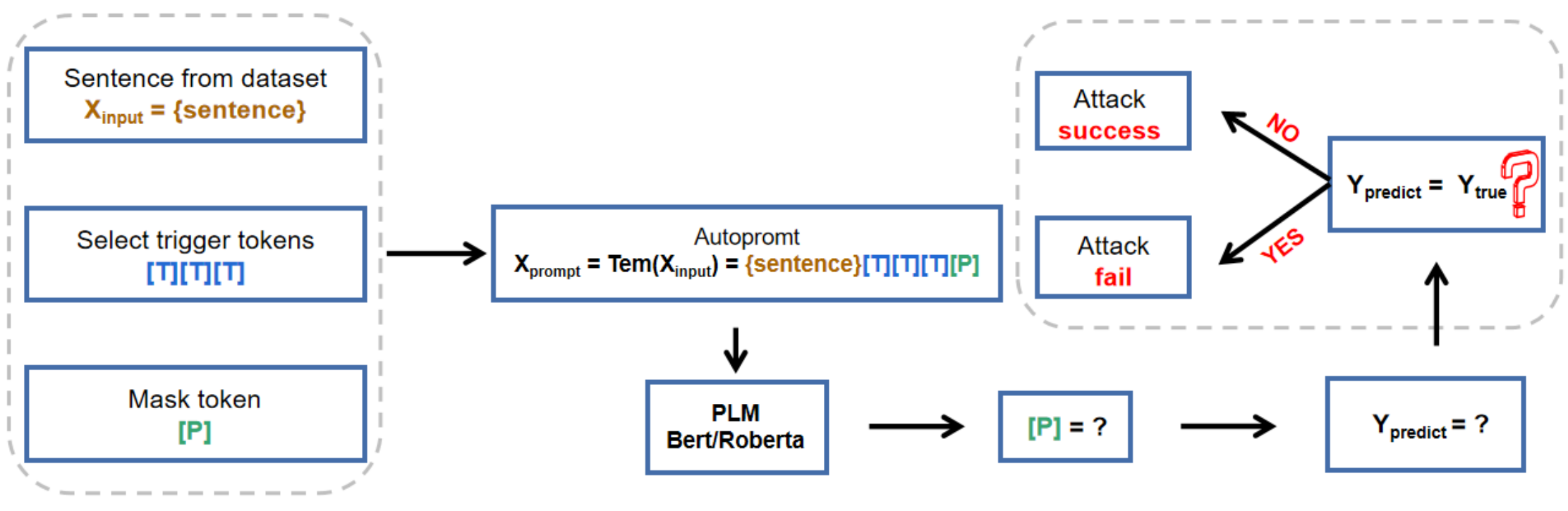}
\caption{The overview of PromptAttack. Firstly, we use the three parts in the virtual box on the left to form a template. Then, we use the template as the input of PLMs, and the output is the predicted token on the mask position. Finally, the model maps the results to the labels and compares the predicted label with the real label.} \label{fig_overview}
%\vspace{-5mm}
\end{figure}

On basis of Shin's~\cite{shin2020autoprompt} Autoprompt method, we propose our attack method PromptAttack. The main idea is to select the tokens that can make the final prediction result drop to form our template. So how to select the final tokens become the most critical problem. Initially, we find trigger tokens as candidate words by using the gradient search method. And then we choose the token sequence with the best attack result as the final template by random replacement or beam search algorithm. The whole process of PromptAttack is shown in Fig.\ref{fig_overview}. 

%\vspace{-0.5cm}

For the convenience of distinction, the sentences obtained from the dataset are denoted by $X_{input}$. The prompts that are input to the PLMs are denoted by $X_{prompt}$. The template defines the format of the prompt: {$X_{input}$} [T] [T] [T] [P] (shown in Fig.\ref{fig_overview}). Among them, [T] represents the trigger token we automatically searched, and [P] is the masked token. Each label corresponds to more than one word, and we denote the set of words for all labels as $V_y$, and then define the verbalizer as an injective function $v$: $L_y$→$V_y$.

\subsection{Select candidate words based on gradient search} We propose an automatic malicious prompt construction method inspired by Wallace et al.~\cite{wallace2019universal}. Our idea is to add some trigger tokens that are shared across all prompts. These tokens are initialized as [MASK] tokens, and our goal is to minimize the predicted probability $\mathcal{P}(y|X_{prompt})$ in Eq(1) of the true label by iterative updating.
\begin{equation}
\mathcal{P}(y|X_{prompt}) = \sum_{w\in V_y} \mathcal{P}([MASK] = w|X_{prompt})
\end{equation}

Use each token (represented by $w$) in the vocabulary $V$ to replace the \textit{j}-th token in the trigger token sequence in turn and then calculate the loss of $w$ on the true label $y$. The first-order approximation of the loss change of this replacement process is expressed as Eq(2):
\begin{equation}
Approximation(w) = \textit{\textbf{w}}_{in}^T \nabla log \mathcal{P}(y|X_{prompt})
\end{equation} where \textit{$\textbf{w}_{in}$} is the input embedding of token \textit{w}. We backpropagate to get the gradient of token \textit{w}, which is $\nabla log \mathcal{P}(y|X_{prompt})$ in Eq(2).

Since we calculate Eq(2) based on the real label, the larger the value, the more likely that the token will cause the final prediction to be correct, and vice versa. To achieve the desired effect of eventually causing the prediction result to be wrong, we choose the \textit{k} words with the smallest $Approximation$ value as the candidate set $V_{cand}$ of the trigger token, as shown in Eq(3):
\begin{equation}
V_{cand} = \mathop{top\text{-}k} \limits_{w\in V} (-Approximation(w))
\end{equation}

\subsection{Selection of token sequences} In the previous section, we introduce how to choose candidate words for trigger tokens, so the next step is how to reasonably choose the token sequences that meet our requirements. The selected sequence will be used as the final template to participate in training and testing. In this part, we propose three methods for selecting token sequences: random replacement, beam search, and GPT-2.

\subsubsection{\textbf{Random replacement strategy.}} In each iteration, we randomly select a position in the token sequence and traverse the tokens in the candidate set to replace the word at that position. Each replacement will form a new prompt to interact with the PLMs. We choose the least accurate sequence on the train dataset as the final template of the current iteration and evaluate its accuracy on the test dataset. After multiple rounds of iterations, we select the one with the lowest accuracy in the testing phase as the ultimate template. The specific algorithm is shown in Algorithm~\ref{alg::replacement}.
\begin{algorithm}[!t]
  \caption{Random replacement}
  \label{alg::replacement}
  \begin{algorithmic}[1]
    \Require
      Trigger token candidates:  $\rm candidates$;
      
      Trigger token number:  $\rm N$
      
      Iteration number:  $\rm K$
    \Ensure
      The token sequence:  $final\_sequence$
     
    \State \rm $Tem = [``[MASK]'',``[MASK]'',``[MASK]'',``[MASK]'',``[MASK]'']$
    
    \State $ M = len(candidates)$
    
    \State $ accuracy[M] = {0}$
        
    \State $best\_metric = accuracy(test\_file, Tem)${\Comment{accuracy on test dataset}}
    
    \State $final\_sequence = Tem$
    
    \For {${i=0}; {i<K}; {i++}$} 
     
    \State $pos = random.randrange(N)$  {\Comment{Random select one position}}
    
    {\Comment{Each word in cnadiates replaces the word at the specified position in turn}}
    \For {$m$ in $enumerate(candidates)$} 
    \State $Tem[pos] = m$
    \State $accuracy[i] = train\_accuracy(train,Tem)${\Comment{accuracy on train dataset}}
    \EndFor
    
    {\Comment{Select the sequence with the best result in this iteration\quad\quad\quad\quad\quad\quad\quad\quad\quad\quad}}
    \State $best\_{candidate\_score} = accuracy.min( )$
    \State $best\_sequence = accuracy.argmin( )$
    \State $test\_metric = accuracy(test\_file, best\_sequence)${\Comment{accuracy on test dataset}}
    \EndFor
    
    \If{$test\_metric < best\_metric$}
    {\Comment{select the best sequence and best metric}}
    
    \State $final\_sequence = best\_sequence$
    
    \State $best\_metric = test\_metric$
    \EndIf
  \end{algorithmic}
\end{algorithm}

\subsubsection{\textbf{Beam search method.}} During the experiment, we find that the random replacement method may not replace the [MASK] token in all positions, especially in the case of few-shot. Therefore, we consider using the beam search strategy, which has better stability. 

\begin{figure}[!t]
\centering
\includegraphics[width=\textwidth]{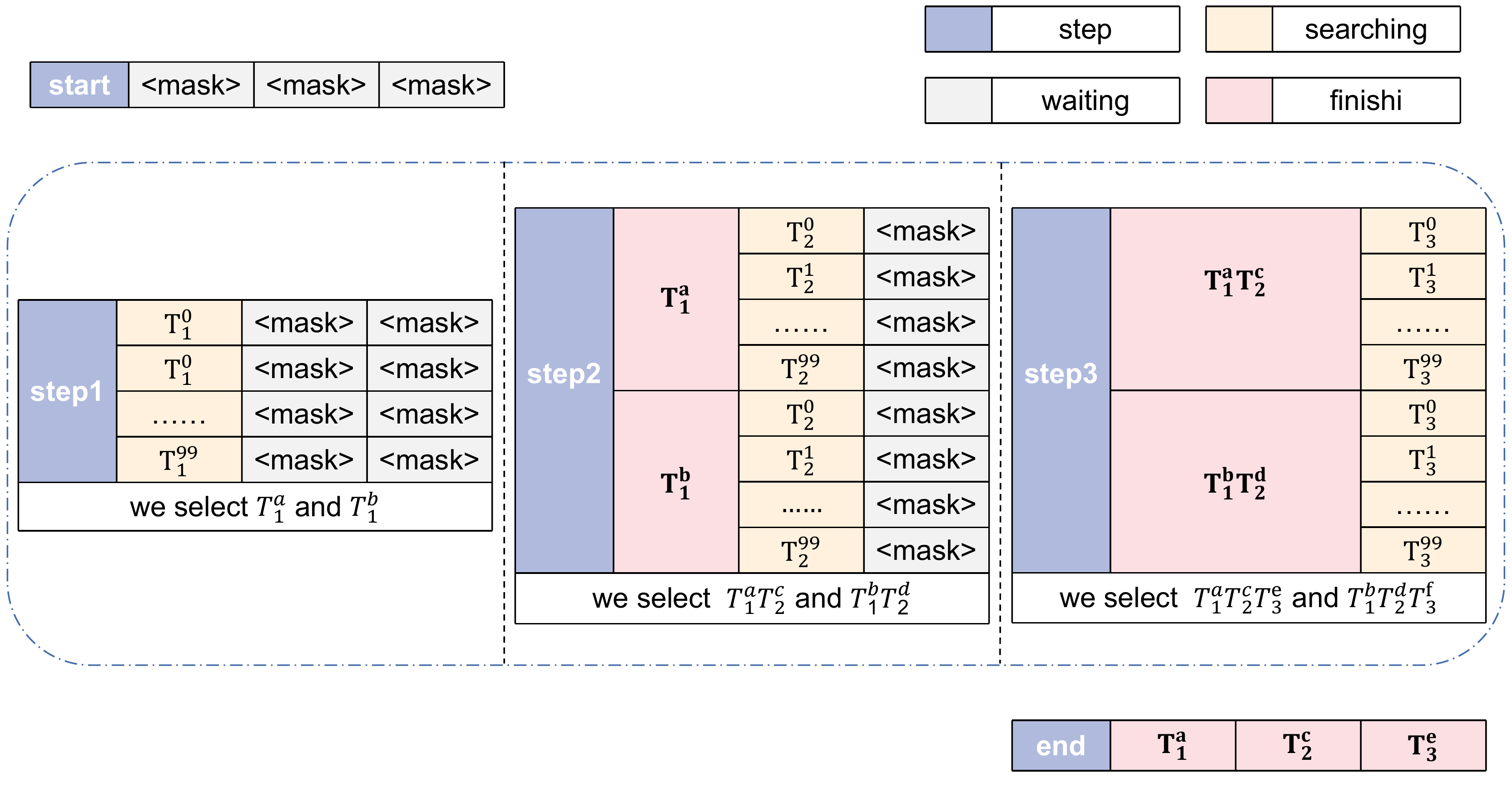}
\caption{The example of token selection for beam search method. In this example, candidate number = 100, beamwidth = 2, token number = 3. After each step, we choose the two templates with the lowest model accuracy as the basis for the next step.} \label{fig2}
\end{figure}

All trigger tokens are initialized to [MASK] and are selected from left to right. Beam search has a hyperparameter beam size, set to $k$. In the first time step, all candidate words replace the <mask> in the first position in turn and are evaluated on the training set. We select the \textit{top-k} sequences, which have the best result on the train dataset, as the candidates at the first step. In the \textit{i}-th time step, based on the sequences of the previous step, all candidate words replace the token in the \textit{i}-th position in turn, and select the \textit{top-k} sequences with the best attack effect as candidate sequences in this step. At last, we select the optimal trigger token. The method is shown in Fig.\ref{fig2}.
%\vspace{-5mm}
\subsubsection{\textbf{Generate token sequences using GPT-2.}} Both the random replacement strategy and the beam search strategy simply focus on how to improve the attack success rate and don't consider the \textbf{fluency of the generated token sequences}. Therefore, we consider using GPT-2 to generate the token sequence. Further, the readability of the template is enhanced, and the approach also increases the concealment of the generated malicious template.

Initially, we train GPT-2 model with all the text in the train dataset. Each token in the candidate set is used as the first token in turn and we use the trained GPT-2 model to generate sequences based on the first word. The sequence with the worst result on the training set has been selected as the final sequence, and its performance on the test dataset and perplexity (PPL) were assessed.

\subsection{Automatic selection of label mapping} We adopt the automatic label selection method proposed by Shin et al.~\cite{shin2020autoprompt}. We select the set of tokens $V_y$ by an automatic method with two steps. 

In the first step, we train a logistic classifier to predict labels using the context containing [MASK] token at \textit{i}-th position. The encoding result by transformers of the context is devoted by $h^{(i)}$. This output $\mathcal{P}(y|h^{(i)})$ (in Eq(5)) can express the relationship of the input and the label. 
\begin{equation}
\mathcal{P}(y|h^{(i)}) \propto exp(h^{(i)}\boldsymbol y+\beta _y),
\end{equation} where $\boldsymbol y$ and $\beta _y$ are the learned weights and bias terms for label $y$, and $i$ denotes the index of the [MASK] token.

In the second step, we replace $h^{(i)}$ with the output word embedding \textit{$\textbf{w}_{out}$} of the PLMs to obtain the score $s(y, w)$. It is known that the greater the correlation between $h^{(i)}$ and $y$, the larger the $\mathcal{P}(y|h^{(i)})$. So the larger the $s$, the more relation between the word and label. So we choose the $k$ highest scoring words to construct the label set $V_y$.

\section{Experiments}
Sentiment analysis is a fundamental task in NLP, which refers to classifying texts into two or more types according to the meaning and emotional information expressed in the text. Our experiments are mainly for the sentiment classification task, and we use the PyTorch implementation and pre-trained weights provided by the transformer Python library~\cite{wolf2019huggingface}.
\subsection{Datasets} We use three sentiment classification datasets, namely \textbf{SST-2}~\cite{socher2013recursive}, \textbf{IMDB}, and \textbf{Amazon Video\_Games}.

\textbf{SST-2 dataset.} SST-2 (The Stanford Sentiment Treebank), a single-sentence classification dataset, contains sentences from movie reviews and human comments on their sentiments. The dataset is divided into two categories based on emotion: positive mood (label corresponds to 1) and negative mood (label corresponds to 0). The SST-2 dataset is from \href{https://github.com/ucinlp/autoprompt}{https://github.com/ucinlp/autoprompt.}.

\textbf{IMDB dataset.} IMDB (Internet Movie Database) is an online database of movie actors, movies, TV shows, TV stars, and movie productions. Labels are represented by 0 and 1, with 0 representing negative and 1 representing positive. 
The IMDB dataset is from \href{https://www.kaggle.com/datasets/uttam94/imdb-mastercsv}{https://www.kaggle.com/datasets/uttam94/imdb-mastercsv}.

\textbf{Amazon Video\_Game dataset.} The dataset we use is the sub-dataset of the Amazon dataset about Video\_Games. We select the summary of users' reviewers as the inputs of the sentiment classification task. In our experiments, we consider reviews rated 1.0 and 2.0 as negative (the label is 0), and reviews rated 4.0 and 5.0 as positive (the label is 1). We get the Amazon review dataset from \href{https://nijianmo.github.io/amazon/index.html}{https://nijianmo.github.io/amazon/index.html}.

\subsection{Setup} In the following sections, we use the Bert\_base model, Roberta\_base model, and Roberta\_large model as pre-trained models respectively.

\textbf{Dataset processing.} On each kind of dataset, we select 24,000 pieces of data as the train dataset, including the same account of data for each category. We choose approximately 1000 pieces of data to form the test dataset, and the data in each category is also equal.

\textbf{Baseline.} We choose the Autoprompt method proposed by Shin et al.~\cite{shin2020autoprompt} as the baseline. And we evaluate the attack performance of PromptAttack by calculating the drop in the model accuracy under the same conditions.

\textbf{Parameter settings.} The number of tokens constituting the template is 5, in other words, the template's format is ``\{sentence\} [T] [T] [T] [T] [T] [P].'' The number of words corresponding to the label is 3. The batch size is 24 in the train process and 48 in the test process. The candidate set includes 100 trigger tokens.
\subsection{Result and analysis} 
\subsubsection{Selection results of words corresponding to labels.} In this process, we perform 50 iterations, and each iteration outputs 50 words that can be mapped with the label. We choose top-3 among them as our final results. Table ~\ref{tab1} shows the label corresponding words on the \textbf{SST-2} dataset under the three pretrained models.

It can be seen from Table~\ref{tab1} that the label words found by different models are completely not the same. Because the word segmentation rules of Bert and Roberta are different, the corresponding words cannot be identical. From the above results, most of the words corresponding to label 0 are derogatory words, and most of the words corresponding to label 1 are commendatory words, which is in line with our cognition of manually selecting label words.
%\vspace{-5mm}
\begin{table}[!t]
\centering
\caption{Results of label mapping.}\label{tab1}
\renewcommand\arraystretch{1.3}
\begin{tabular}{|m{2.1cm}<{\centering} |m{4.9cm}<{\centering} | m{4.9cm}<{\centering}|}
\hline 
    &  Labe : 0 & Label : 1 \\
\hline
Bert\_base & \{coward, \#\#cket, Ordnance\} & \{extraordinary, wealth, natural\} \\
\hline
Roberta\_base & \{worst, ĠWorse, Ġblames\} & \{illance, ĠLens, shine\} \\
\hline
Roberta\_large & \{Ġworse, Ġincompetence, ĠWorse\} & \{ĠCris, Ġmarvelous, Ġphilanthrop\}\\
\hline
\end{tabular}
\vspace{-5mm}
\end{table}
%\vspace{-8mm}

\subsubsection{Experimental results of the \textbf{SST-2} dataset on three pre-trained models.}
The data outside the parentheses are the model prediction accuracy(Accuracy). The part in parentheses is the percentage of accuracy drop, that is, the difference(Diff) between PromptAtatck's accuracy and the Autoprompt's accuracy; the bolded part represents the best attack effect of the \textbf{SST-2} dataset under each model.

It can be seen from the data in Table~\ref{tab2} that our two methods can effectively reduce the accuracy no matter which pre-trained model is based on. Among them, the beam search method based on the Roberta\_large model achieves the best attack effect. The attack success rates of the two strategies on Robrta\_large can respectively reach 47.36\% and 56.77\%. In addition, we can see that the attack effect based on the Roberta model is better than the result based on the Bert model. 

%\vspace{-5mm}
\begin{table}
\centering
\caption{Results of the SST-2 dataset on three pre-trained models. Metric : Accuracy (Diff)}\label{tab2}
\setlength{\tabcolsep}{2mm}
\renewcommand\arraystretch{1.3}
\begin{tabular}{m{2cm}<{\centering}|m{2cm}<{\centering}|m{3cm}<{\centering}|m{3cm}<{\centering}}
\hline 
    &  Autoprompt & Random & Beamsearch \\
\hline
Bert & 82.30\% & 44.61\%\quad\textbf{(37.69\%)} & 49.08\%\quad(33.22\%) \\
\hline
Roberta\_base & 85.44\% & 46.90\%\quad(38.54\%) & 44.03\%\quad\textbf{(41.41\%)} \\
\hline
Roberta\_large & 89.79\%	& 42.43\%\quad(47.36\%) & 33.02\%\quad\textbf{(56.77\%)} \\
\hline
\end{tabular}
\vspace{-8mm}
\end{table}

\subsubsection{Results of \textbf{SST-2}/\textbf{IMDB}/\textbf{Amazon} dataset on Roberta\_large.}From Table~\ref{tab2} we can see that the \textbf{SST-2} dataset performs best on Roberta\_large model, so we evaluate the \textbf{IMDB} dataset and the \textbf{Amazon} dataset on Roberta\_large model. The experimental results are shown in Table~\ref{tab3}.

The data in Table \ref{tab3} shows that based on the Roberta\_large model, PromptAttack is effective on all three datasets, and the attack success rate can reach more than 40\%. The results also verify the general applicability of our method.

%\vspace{-5mm}
\begin{table}[!t]
\centering
\caption{Results of \textbf{SST-2}/\textbf{IMDB}/\textbf{Amazon} dataset on Roberta\_large. Metric : Accuracy (Diff)}\label{tab3}
\setlength{\tabcolsep}{2mm}
\renewcommand\arraystretch{1.3}
\begin{tabular}{m{2cm}<{\centering}|m{2cm}<{\centering}|m{3cm}<{\centering}|m{3cm}<{\centering}}
\hline
    &  Autoprompt & Random & Beamsearch \\
\hline
\textbf{SST-2} & 89.79\%	& 42.43\%\quad(47.36\%) & 33.02\%\quad\textbf{(56.77\%)}  \\
\hline
\textbf{IMDB} & 84.45\% & 43.23\%\quad\textbf{(41.22\%)} & 49.69\%\quad(34.76\%) \\
\hline
\textbf{Amazon} & 87.23\% & 42.00\%\quad\textbf{(45.23\%)} & 46.9\%\quad(40.33\%) \\
\hline
\end{tabular}
%\vspace{-5mm}
\end{table}
%\vspace{-8mm}

\subsubsection{Validation of the method of GPT-2.}Considering the invisibility of automatically constructing templates, we propose the method of using GPT-2 to generate smoother templates. The readability of the template is evaluated using PPL (perplexity), and the experimental results based on the Roberta\_large model are shown in Table~\ref{tab4}.

The table shows three indicators: perplexity (PPL), model prediction accuracy, and the generated token sequences. From Table~\ref{tab4}, we can first see that the token sequence generated by the GPT-2 method can reduce the accuracy of the model, that is, the attack can be successful. It also can be seen that the token sequence generated by the GPT-2 method has the lowest perplexity, which proves the effectiveness of the GPT-2 method. At the same time, we can find that the beam search method has the lowest accuracy and the highest perplexity, while the GPT-2 method has the highest accuracy and the lowest perplexity. That is to say, while the perplexity of the generated token sequence is reduced, it will also lead to the deterioration of the experimental attack performance.
%\vspace{-5mm}
\begin{table}
\centering
\caption{The perplexity of the template in different ways.}\label{tab4}
\setlength{\tabcolsep}{2mm}
\renewcommand\arraystretch{1.3}
\begin{tabular}{m{2cm}<{\centering}|m{1cm}<{\centering}|m{1.5cm}<{\centering}|m{5cm}<{\centering}}
\hline
    & PPL & Accuracy & Token sequence \\
\hline
Beamsearch & 1506.17 & 33.02\% & Ġwhereas ĠHammer ` ' Ġeffectiveness Ctrl \\
\hline
Random & 144.94 & 42.43\% & ĠCue Ġimperfect Ġpackets Ġperfume ĠNer\\
\hline
GPT-2 & 76.68 & 49.08\% 
 & Ġconvenience, Ġseries Ġdisgusting Ġunfortunate Ġdisable
 \\
\hline
\end{tabular}
\vspace{-8mm}
\end{table}

\subsubsection{Experimental results in the case of few-shot.}The fact that the prompt is also valid in few-shot situations is an important advantage of the method. The baseline method has verified in their paper that the Autoprompt method is effective in the few-shot case. Therefore we try to explore whether PromptAttack is useful in the few-shot case.

We randomly select 100 pieces from each category of data in each dataset to form a few-shot dataset and conduct attack experiments on Roberta\_base and Roberta\_large models. And we obtain the following experimental results, as shown in Table \ref{tab5}.

We can see that all three datasets can be successfully attacked in the few-shot case on the two pre-trained models. It is worth noting that on the \textbf{SST-2} dataset and \textbf{Amazon} dataset, the attack success rate of Roberta\_base is higher than that of Roberta\_large. In our analysis, the reason for this may be that most of the texts in these two datasets are short sentences. When the train data is small and the text length is short, the larger-scale model will overfit in classification. The text of the \textbf{IMDB} dataset is mostly long sentences, which is not easy to overfit, so the attack success rate on the Roberta\_large model is still higher.

%\vspace{-5mm}
\begin{table}[!t]
\centering
\caption{Results under few-shot conditions. Metric : Accuracy (Diff)}\label{tab5}
\setlength{\tabcolsep}{1.5mm}
\renewcommand\arraystretch{1.3}
\begin{tabular}{m{1.2cm}<{\centering}|m{2.0cm}<{\centering}|m{1.7cm}<{\centering}|m{2.8cm}<{\centering}|m{2.8cm}<{\centering}}
\hline
Dataset & Model & Autoprompt & Random & Beamsearch \\
\hline
\multirow{2}*{\textbf{SST-2}} & Roberta\_base & 80.16\% & 49.08\%\quad(31.08\%) & 40.48\%\quad\textbf{(39.68\%)}\\
\cline{2-5}
~ & Roberta\_large & 73.97\% & 46.69\%\quad(27.28\%) & 46.22\%\quad\textbf{(27.75\%)} \\
\hline
\multirow{2}*{\textbf{Amazon}} & Roberta\_base & 74.70\% & 47.01\%\quad\textbf{(27.69\%)} & 48.05\%\quad(26.65\%)\\
\cline{2-5}
~ & Roberta\_large & 70.69\% & 43.47\%\quad\textbf{(27.22\%)} & 45.70\%\quad(24.99\%) \\
\hline
\multirow{2}*{\textbf{IMDB}} & Roberta\_base & 66.46\% & 49.27\%\quad(17.19\%) & 44.79\%\quad\textbf{(21.67\%)}\\
\cline{2-5}
~ & Roberta\_large & 69.98\% & 46.67\%\quad\textbf{(23.31\%)} & 48.85\%\quad(21.13\%) \\
\hline
\end{tabular}
\end{table}
%\vspace{-8mm}
\section{Conclusion and future work} Most current researches ignore the security issues of the prompt model, and especially its unique template part is vulnerable to attack. Therefore, we propose an attack method PromptAttack based on the template innovatively. The method is evaluated by three datasets for sentiment classification and three pre-trained language models. The experiment results verify the effectiveness of our attack way and show that our method is also applicable to the few-shot case.

In the future, we can continue to research prompt security issues. On the one hand, PromptAttack only targets discrete prompts, and the follow-up study can focus on attacking continuous prompts. On the other hand, our method is a white-box attack that needs to know all the parameters of the model, and we can try to design a black-box attack method.

%\section{Acknowledgements}

\bibliographystyle{splncs04}
\bibliography{reference.bib}
\end{document}